\documentclass[lettersize,journal]{IEEEtran}
\usepackage{amsmath,amsfonts}
\usepackage{algorithmic}
\usepackage{algorithm}
\usepackage{array}
\usepackage{amssymb}
\usepackage[caption=false,font=normalsize,labelfont=sf,textfont=sf]{subfig}
\usepackage{textcomp}
\usepackage{stfloats}
\usepackage{url}
\usepackage{verbatim}
\usepackage{graphicx}
\usepackage{cite}
\usepackage{multirow}
\usepackage{float}
\usepackage{hyperref}

\begin{document}

\title{Boosting the Class-Incremental Learning in 3D Point Clouds via Zero-Collection-Cost Basic Shape Pre-Training}

\author{Chao~Qi,
        Jianqin~Yin,
        Meng~Chen,
        Yingchun~Niu,
        and~Yuan~Sun
\thanks{This work was supported by the National Natural Science Foundation of China under Grant 62403491 \emph{(Corresponding author: Jianqin Yin.)}

Chao Qi, Jianqin Yin, Meng Chen, and Yingchun Niu are with the School of Intelligent Engineering and Automation, Beijing
University of Posts and Telecommunications, Beijing 102206, China. (e-mail: qichao199@163.com; jqyin@bupt.edu.cn; chenmeng@bupt.edu.cn; ycniu@bupt.edu.cn)

Yuan Sun is with the School of Electronic Engineering, Beijing
University of Posts and Telecommunications, Beijing 100876, China. (e-mail: sunyuan@bupt.edu.cn)
        }%
}

\markboth{Journal of \LaTeX\ Class Files,~Vol.~14, No.~8, August~2021}%
{Shell \MakeLowercase{\textit{et al.}}: A Sample Article Using IEEEtran.cls for IEEE Journals}

\maketitle

\begin{abstract}
Existing class-incremental learning methods in 3D point clouds rely on exemplars (samples of former classes) to resist the catastrophic forgetting of models, and exemplar-free settings will greatly degrade the performance. For exemplar-free incremental learning, the pre-trained model methods have achieved state-of-the-art results in 2D domains. However, these methods cannot be migrated to the 3D domains due to the limited pre-training datasets and insufficient focus on fine-grained geometric details. This paper breaks through these limitations, proposing a basic shape dataset with zero collection cost for model pre-training. It helps a model obtain extensive knowledge of 3D geometries. Based on this, we propose a framework embedded with 3D geometry knowledge for incremental learning in point clouds, compatible with exemplar-free (-based) settings. In the incremental stage, the geometry knowledge is extended to represent objects in point clouds. The class prototype is calculated by regularizing the data representation with the same category and is kept adjusting in the learning process. It helps the model remember the shape features of different categories. Experiments show that our method outperforms other baseline methods by a large margin on various benchmark datasets, considering both exemplar-free (-based) settings. The dataset and code are available at \href{https://github.com/chaoqi7/BSA-CIL-3D}{https://github.com/chaoqi7/BSA-CIL-3D}.
\end{abstract}


\section{Introduction}
\IEEEPARstart{I}{n} the open world, an agent with a 3D laser scanner observes novel objects in point clouds, assigning objects with different class labels. However, the label space keeps growing. The newly learned class knowledge overwrites previous ones by updating the parameters of computer vision models, causing catastrophic forgetting of former classes. It is the main challenge of class-incremental learning (CIL) in 3D point clouds (-3D).

Some studies have been carried out in CIL-3D. The geometric-aware \cite{RN414, RN466}, knowledge distillation \cite{RN410}, and regularization \cite{RN413} methods cooperate with the data replay mechanism \cite{RN452}, resisting forgetting previously learned class categories well. However, even in different ways, all these methods rely on exemplars to remember former classes. \cite{RN411} proposed an exemplar-free point cloud class incremental benchmark. However, the performance deteriorates significantly in the later incremental stages.

\begin{figure}
    \centering
    \includegraphics[width=1\linewidth]{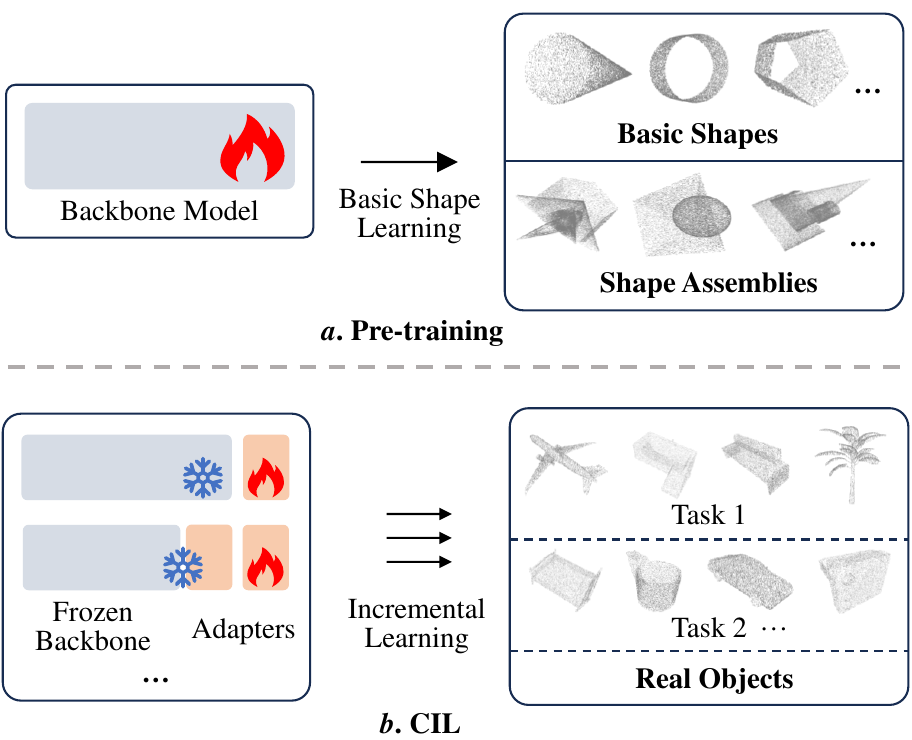}
    \caption{In the pre-training stage, the backbone model learns knowledge from the zero-collection-cost basic shapes and shape assemblies. In CIL, the backbone is frozen to remember geometry knowledge, introducing adapters to incremental learn real objects in point clouds.}
    \label{fig1}
\end{figure}

Exemplar-free (-based) CIL methods were widely discussed in 2D domains and inspired CIL-3D a lot. Recently, the pre-trained model-based methods have demonstrated state-of-the-art (SOTA) performance in CIL of image classifications \cite{RN465, RNA460, RN421, RN617}. By pre-training the model on a large-scale dataset (e.g., ImageNet \cite{RN425} with millions of samples covering thousands of class categories), the models obtain extensive knowledge of 2D features \cite{RN421}. After that, adapter layers are introduced into the model, cooperating with the generalizable knowledge to learn subsets of CIL benchmarks (e.g., Omni-benchmark \cite{RN426} with \textit{a large domain gap} with the pre-training dataset) continually. It prevents the model from forgetting former classes effectively.

In early research, the pre-trained model methods in CIL of 3D point clouds are limited by the datasets. Specifically, there are relatively few datasets for point cloud classification, among which ShapeNet55 \cite{RN423} and ModelNet40 \cite{RN424} are the most widely used. The largest ShapeNet55 only contains thousands of objects, covering 55 class categories. Pre-training on ShapeNet55 results in a model with biased knowledge due to the limited data samples. Besides, More than half of the categories of ShapeNet55 are similar to those of ModelNet40. If we use ModelNet40 for CIL, most incremental learning class categories have been seen in the pre-training stage. The \textit{small domain gaps} between pre-training and CIL datasets can not verify the incremental class learning ability.

Recently, large language models (LLM) for 3D points, such as the PointLLM \cite{RN601}, have received attention. It processes object point clouds with human instructions and provides extensive knowledge for the 3D vision model. However, LLM's focus on global feature extraction may overlook fine-grained local geometric details, which are crucial for distinguishing between similar point cloud categories in CIL.

In this paper, we propose an easy and interesting method to address the above problems. In the 3D world, an object is composed of some basic 3D shapes \cite{RN604, RN605}. Creating a 3D object begins with recognizing basic shapes and their assemblies, which can then be combined into more complex forms \cite{RN609}. For example, four cylinders and one polyhedron can assemble a table. Marr \cite{RN607} introduced this concept from the field of cognitive science into the domain of computer vision, stating that the visual systems of humans (as well as artificial intelligence) decompose objects into basic shapes in order to recognize 3D forms continually. Inspired by this, we aim to enhance the vision model's ability to recognize basic shapes as the foundation for CIL-3D.

Specifically, we propose a novel dataset, including \textbf{B}asic \textbf{S}hapes (cone, cylinder, ellipsoid, polyhedron, prism, pyramid) and basic shape \textbf{A}ssemblies (namely \textbf{BSA}). This dataset, generated through shape formulas and specific rules, is cost-free and rich in fine-grained geometric details. Pre-training on the basic shapes helps a model obtain extensive knowledge of 3D geometry features, as shown in figure \ref{fig1}. Besides, the shape assembly samples are semantic-agnostic, naturally leaving a large domain gap with the CIL ones.

To reduce the model catastrophic forgetting in the incremental learning stage, we propose a CIL-3D framework compatible with exemplar-free (-based) settings. In every CIL task, we introduce novel adapters to the model and extend the geometry knowledge to represent the data samples. Exemplar-adaptive regularizations prompt the data sample representations of the same category to be similar, forming the class prototype (a template in the embedding space \cite{RN449}) with or without exemplars. Every class prototype is calculated using the same basic geometry knowledge and adjusted with the learning process. It helps the model remember the observed object features and reduces its catastrophic forgetting. Our contributions can be summarized as:

\begin{itemize}
    \item We propose a zero-collection-cost pre-training dataset BSA. It consists of basic shapes (assemblies) and provides extensive geometry knowledge to the model, significantly enhancing its incremental learning capabilities.
    \item We propose a framework infused with the geometry knowledge for CIL-3D. It regularizes data representation and dynamically adjusts class prototypes along the incremental stage, effectively mitigating catastrophic forgetting of models with or without exemplars.
    \item We conduct experiments with different exemplar settings, achieving SOTA results on all the real-world and synthetic benchmark datasets. Our method outperforms the baseline methods by a large margin. Besides, the basic shape pre-training manner far surpasses the LLMs in CIL-3D.  
\end{itemize}

\section{Related Works}

\subsection{Class-Incremental Learning}

Many works have explored methods to address the CIL problems, which can be divided into the following categories \cite{RN449}. The data replay methods \cite{RN459, RN429, RN451, RN450, RN620, RN622} select exemplars from former classes and recover the prior knowledge while learning novel ones. These exemplar-based methods are widely used and cooperate well with other methods \cite{RN437, RN416, RN438, RN446}. For example, knowledge distillation methods \cite{RN439, RN440, RN441, RN442, RN443} use the teach-student framework to distill prior knowledge from exemplars to reduce model catastrophic forgetting. Besides, some works have proposed regularization methods \cite{RN430, RN436} and rectification methods \cite{RN446, RN444, RN445, RN447} to minimize model forgetting, either by regularizing parameters or lowering the biased prediction of models.

Recently, dynamic network methods \cite{RN417, RN431, RN432, RNA461, RN434, RN621} have proven their effectiveness in CIL. It improves the ability to represent incremental class categories by adjusting network structures. As a dynamic network variant, a pre-trained Vision Transformer (ViT) with learnable increasing layers has demonstrated excellent performance in 2D domains \cite{RN421, RN448}. However, it is limited in CIL-3D due to the small datasets and the similar class categories between the pre-training and CIL datasets. It inspires us to find a way to address the problem.

\subsection{Class-Incremental Learning in 3D Point Clouds}

CIL-3D has received attention recently due to its potential applications in robotics, autonomous driving, and augmented reality. Dong et al. \cite{RN414, RN466} pioneered the exploration of CIL-3D and designed a geometric-aware attention mechanism to prevent the catastrophic forgetting brought by redundant geometric information in point clouds. This approach selectively focuses on 3D features, ensuring that critical geometry details are preserved during incremental learning. Building on this, \cite{RN410} used the knowledge distillation method to transfer and update the shared 3D point knowledge in the incremental learning process, leveraging the teacher-student framework to maintain prior knowledge while assimilating new information. \cite{RN413} used point cloud rehearsal and reconstruction as regularization methods, significantly decreasing catastrophic forgetting in the learning process. All these methods heavily depend on exemplars to remember former knowledge. However, due to memory constraints or data legality issues (e.g., privacy concerns, data ownership), exemplars may not always be accessible. This has spurred the need for exemplar-free CIL-3D approaches to achieve incremental learning without retaining past data samples.

\cite{RN411} proposed an exemplar-free CIL-3D benchmark, addressing this gap by proposing a framework that operates entirely without exemplars. However, the performance deteriorates significantly in the later CIL stages. It motivates us to explore a CIL-3D framework that can work well with exemplar-based and exemplar-free settings.

\begin{figure*}
    \centering
    \includegraphics[width=0.8\linewidth]{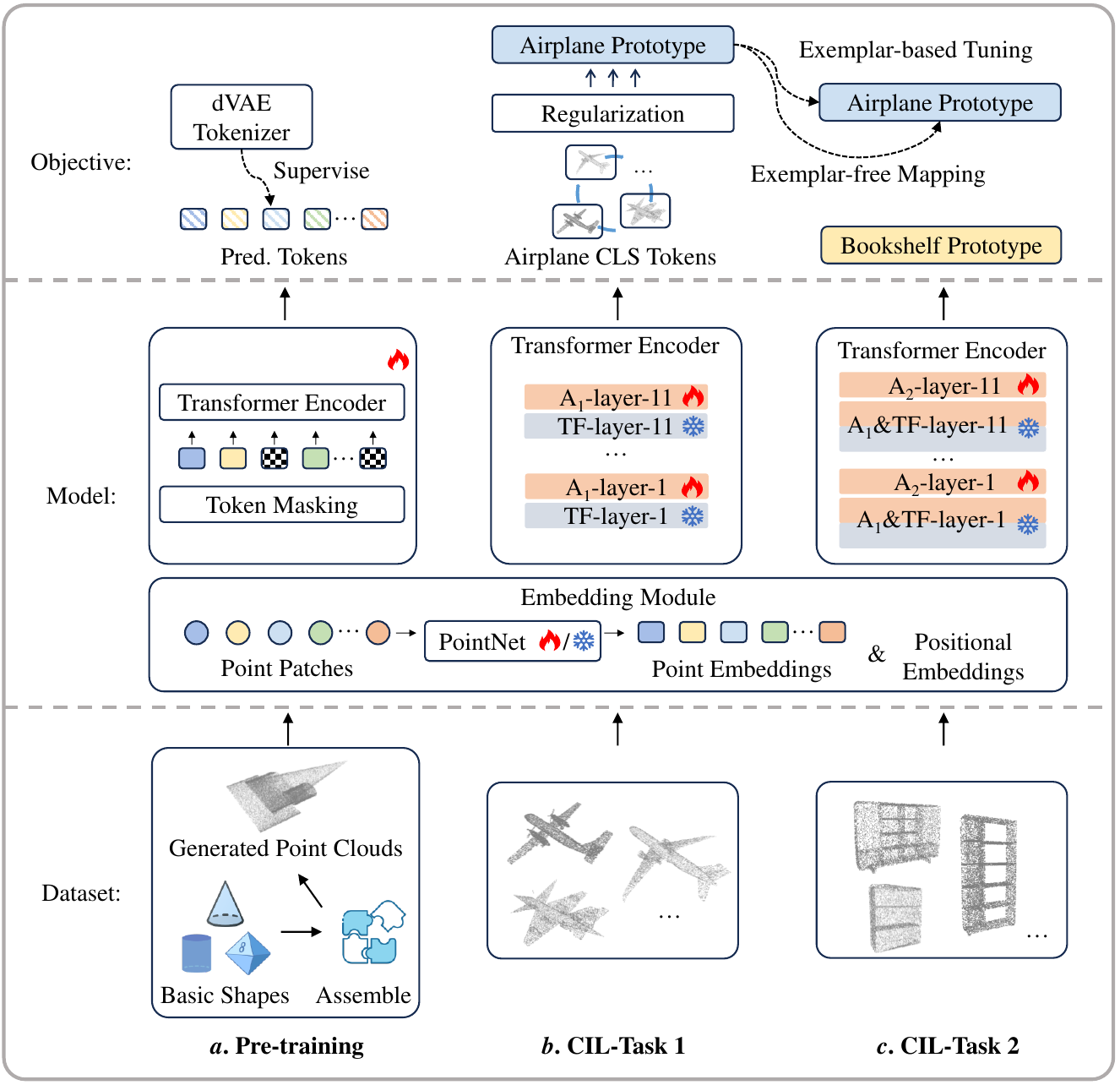}
    \caption{The pre-training dataset is generated by shape formulas and specific regulars without any collection cost. A discrete Variational AutoEncoder (dVAE) supervises the predicted tokens of the pre-training dataset. In CIL-task \textit{t}, the transformer encoder introduces adapter layers to output [CLS] Tokens, cooperating with an exemplar-adaptive regularization method to calculate class prototypes, where A${_t}$-layer-\textit{n} indicates the adapter layer following the \textit{n}-th transformer (TF-) layer. In task \textit{t}+1, TF- and A${_{t-1}}$-layers are frozen, and the learnable A${_t}$-layers update former class prototypes by exemplar-based tuning or exemplar-free mapping method.}
    \label{fig2}
\end{figure*}

\section{Problem Statement}

The class-incremental learning in 3D point clouds can be formulated as follows: Given a sequence of \textit{T} training sets $D = \left\{ {{D_1},{D_2}, \ldots ,{D_T}} \right\}$ in point clouds, where ${D_t} = \left\{ {\left( {{x_i},{y_i}} \right)} \right\}_{i = 1}^{{n_t}}$ denotes ${n_t}$ data samples in the \textit{t}-th training set. For exemplar-free CIL-3D, the exemplar set $\xi$ is empty; For exemplar-based CIL-3D, $\xi  = \left\{ {{\xi _1},{\xi _2}, \ldots ,{\xi _T}} \right\}$ is the exemplar set and meets the following condition: ${\xi _t} \subset {D_t}$.

Every data sample ${x_i} \in {\mathbb{R}^{m \times c}}$ includes \textit{m} points with \textit{c} channels, and ${y_i} \in {\mathcal{Y}_t}$ indicates the corresponding class label. ${\mathcal{Y}_t}$ is the label space of task \textit{t}, and different tasks have non-overlapping spaces. A CIL-3D model $f( \cdot )$ inputs the training sets in sequence and predicts the label $f({x_i})$ of every data sample.

In task \textit{t}, the model $f( \cdot )$ learns ${D_t}$ with label space ${\mathcal{Y}_t}$. The parameters keep updating, which leads the model to forget the previously learned sets ${D_1} \cup  \ldots {D_{t - 1}}$  with label spaces ${\mathcal{Y}_1} \cup  \ldots {\mathcal{Y}_{t - 1}}$, which is called catastrophic forgetting. We aim to train a CIL-3D model not only to learn ${D_t}$ well, but also not to forget previous ${D_1} \cup  \ldots {D_{t - 1}}
$. It can be denoted as:

\[{f^*} = \mathop {\arg \min }\limits_{({x_i},{y_i}) \in {D_1} \cup  \ldots {D_t}} I({y_i} \ne f({x_i}))\]
where $I( \cdot )$ is the indicator function. For exemplar-based CIL-3D, ${\xi _1} \cup  \ldots {\xi _{t - 1}}$ can be observed in task \textit{t}. However, no samples in the previous sets can be used for the exemplar-free one.

\section{Method}

\subsection{Overview}

We employ a novel CIL-3D framework, as shown in figure \ref{fig2}. Firstly, we propose a pre-training dataset by assembling basic shapes, helping the backbone model learn geometry knowledge. Then, in the incremental learning stage, the growing label space is built upon the learned knowledge. Modules are designed to calculate class prototypes, reducing the model forgetting of former classes considering different exemplar settings.

\subsection{Basic Shapes}

We chose cone, cylinder, ellipsoid, polyhedron, prism, and pyramid as \textit{basic shape elements} \cite{RN604}. \textbf{The supplementary material section \textit{\uppercase\expandafter{\romannumeral 1}} provides more details.}

Shape formulas can create basic shapes in point clouds. Taking the cylinder point cloud as an example, $x_i^j = (r\cos ({\theta ^j}),r\sin ({\theta ^j}),{h^j})$ denotes the \textit{j}-th point in cylinder \textit{i}, where ${\theta ^j}$ is the polar angle and ${h^j} \in [ - 0.5H,0.5H]$ is the point height. Massive points ${x_i} = \{ x_i^j\} _{j = 1}^m$ form a point-cloud-based cylinder. Radius \textit{r} and height \textit{H} decide the cylinder size. Basic shapes in point clouds with different sizes are summarized into a \textit{basic shape pool}. 

\begin{figure}
    \centering
    \includegraphics[width=0.85\linewidth]{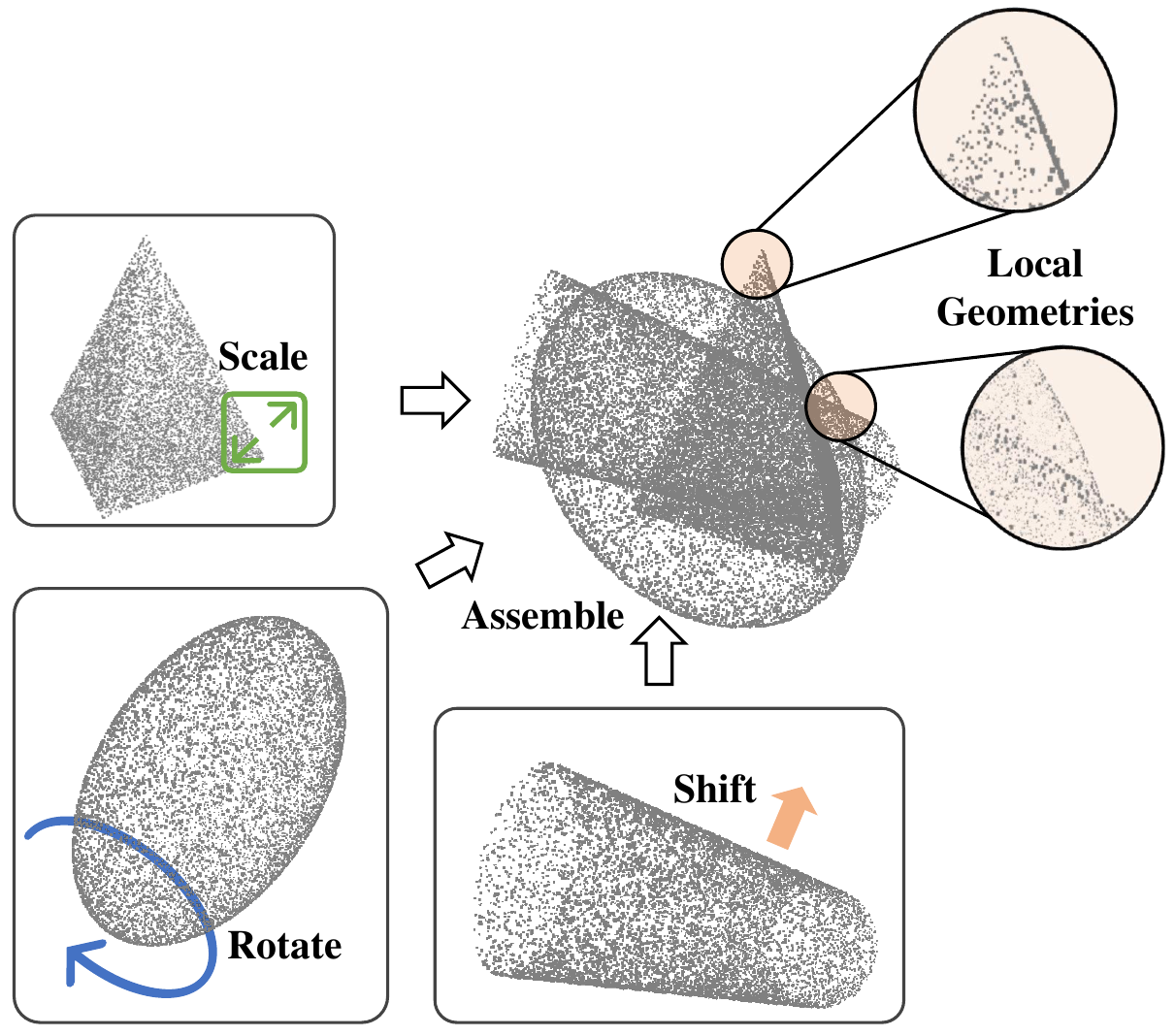}
    \caption{An example of shape assembly. Several basic shapes make a 3D object of our dataset in the point cloud. The object is semantic-agnostic and contains local geometries, which are very similar to those in real objects.}
    \label{fig3}
\end{figure}

A pre-training dataset should contain extensive knowledge of 3D shapes. Thus, we shift, rotate, and scale samples in the shape pool, assembling transformed samples into complex 3D objects in point clouds. These 3D objects contain abundant, meaningful local geometries. An example is shown in figure \ref{fig3}. 

Besides, we classify assembling 3D objects with the same elements and transformation rules into the same category. The category labels do not have realistic meaning and vary greatly from the CIL dataset (ModelNet40, ShapeNet55, etc.) labels. Thus, the pre-training dataset keeps a large gap with the CIL ones. The objects assembled with the samples in the shape pool form the \textit{basic shape dataset} BSA.

\subsection{Model Pre-training with Basic Shapes}

BERT \cite{RN453} achieves impressive performance in the context prediction of language, and the Masked Language Modeling (MLM) strategy inspires lots of pre-training works, such as BEiT \cite{RN454} and Point-BERT \cite{RN455}. We define the BERT-style model as $f( \cdot ) = {W^T}\phi (e( \cdot ))$, where $e( \cdot )$ is an embedding module, $\phi ( \cdot )$ is a transformer encoder, and \textit{W} is a linear classifier.

Like the Point-BERT pre-training for 3D point clouds, we divide the \textit{i}-th 3D object ${x_i}$ into \textit{g} point patches $\{ x_i^j\} _{j = 1}^g$, and a PointNet \cite{RN46} projects the patches into \textit{g} embeddings $\{ e_i^j\} _{j = 1}^g$. A masking function works on the embeddings $\mathcal{M}(\{ e_i^j\} _{j = 1}^g)$, together with the position embedding $\{ pos_i^j\} _{j = 1}^g$ as the input of the transformer encoder $\phi ( \cdot )$. The pre-trained dVAE tokenizer \cite{RN469} encodes local patches into informative point tokens $\{ \tilde e_i^j\} _{j = 1}^g$, which supervise the transformer encoder pre-training.

In this process, the PointNet and transformer encoder are trainable. These modules work together to model the geometric patterns of basic 3D shapes, learning abundant geometry knowledge.

\subsection{Class Prototype Calculations with Exemplar-adaptive Regularizations}

To expand geometry knowledge to recognize a novel object ${x_i}$, we introduce adapter layers A${_t}$ for task \textit{t} \cite{RN458} into the transformer encoder $\phi ( \cdot )$ to learn the object embeddings $e({x_i}) = \{ e_i^j,pos_i^j\} _{j = 1}^g$, denoting as $\phi (e({x_i});{{\rm{A}}_{t}})$. Only a few trainable parameters are added per CIL task to adapt novel objects, while the PointNet and transformer (TF-) layers are frozen to revisit previously learned geometry knowledge.

The geometry features are different even for objects with the same class label. Memorizing one prototype in the deep space for each class \cite{RN459} and adjusting it along the incremental stage effectively resist the model's catastrophic forgetting.

We use the mean [CLS] tokens with exemplar-adaptive ranges to represent the class prototype, which can be denoted as:

\begin{equation}
    {p_{t,{\hat y}}} = \mathbb{E}_{({x_i},{y_i}) \sim {(\tilde D \cup \xi )_{{y_i} = \hat y}}} {[\phi _{[{\rm{CLS}}]}^t({x_i})]}
\end{equation}
where $\xi$ is the exemplar set and $\phi _{[{\rm{CLS}}]}^t({x_i})$ is short for ${\phi _{[{\rm{CLS}}]}}(e({x_i});{{\rm{A}}_t})$. For the exemplar-based CIL, ${\tilde D_{{y_i} = \hat y}}$ is null, and only exemplars are used to calculate the prototypes. For the exemplar-free one, ${\tilde D_{{y_i} = \hat y}}$ equals to all the training samples ${D_{{y_i} = \hat y}}$ with label $\hat y$. While the model observes ${D_{{y_i} = \hat y}}$, a regularization item adapting to different exemplar settings minimizes the representation gap between all the training samples and the class prototype:

\begin{equation}
\mathop {\min }\limits_{({x_i},{y_i})\sim{D_{{y_i} = \hat y}}} [\phi _{[{\rm{CLS}}]}^t({x_i}) - {p_{t,{\hat y}}}]
\end{equation}
In task \textit{t}+1, the model recognizes novel objects and flushes the prototype of former classes. We introduce a new adapter layer  ${{\rm{A}}_{t + 1}}$ and recalculate the former prototypes. If exemplars ${\xi _{{y_i} = \hat y}}$ are left in the memory, the prototypes are tuned based on the exemplars:

\begin{equation}
{p_{t+1,{\hat y}}} = \mathbb{E}_{({x_i},{y_i}) \sim {\xi_{{y_i} = \hat y}}} {[\phi _{[{\rm{CLS}}]}^{t + 1}({x_i})]}
\end{equation}
If no exemplars exist in the memory, we adjust previous class prototypes via semantic mapping \cite{RN421}. It synthesizes new features for old classes, ensuring compatibility without old class instances. Thus, we employ it as one component in our framework's exemplar-free section. Based on this, the exemplar-adaptive regularizations are iteratively enforced.

\subsection{Loss Function}

In task \textit{t} of CIL-3D, our model observes ${D_t}$ and predicts the label of ${x_i}$. Thus, we calculate the cross entropy between the prediction and the true label ${y_i}$. The cross-entropy item in the loss function is illustrated as: $\mathcal{L}_{({x_i},{y_i}) \sim {D_t}}^1 = {\rm{CE}}({W^T}\phi _{[{\rm{CLS}}]}^t({x_i}),{y_i})$

Besides, we aim to reduce the difference between the data representation $\phi _{[{\rm{CLS}}]}^t({x_i})$ and the corresponding class prototype ${p_{t,{y_i}}}$: $\mathcal{L}_{({x_i},{y_i}) \sim ({D_t} \cup {\xi _{1:t - 1}})}^2 = \mathbb{E}{[\phi _{[{\rm{CLS}}]}^t({x_i}) - {p_{t,{y_i}}}]^2}$. Note that, for the exemplar-based setting, the losses brought by the exemplar set ${\xi _{1:t - 1}}$ are considered. The total loss is defined as $\mathcal{L} = {\mathcal{L}^1} + {\mathcal{L}^2}$.

\section{Experiments}

We conduct experiments on benchmark datasets, evaluating the effectiveness of our method on CIL in 3D point clouds. Experiments are designed to answer the following questions: (1) \textit{Can our method effectively resist the model's catastrophic forgetting in the incremental stage}? (2)\textit{ Does our method perform well in exemplar-based and exemplar-free experiment settings}? (3) \textit{Does our proposed basic shape dataset play a significant role in incremental learning of 3D shapes}? (4) \textit{How do several important designs affect the experiment results}?

Sections \textit{\uppercase\expandafter{\romannumeral 5}}-\textit{A} to \textit{D} introduce our experiment's dataset, baselines, implementation details, and evaluation metrics. Sections \textit{\uppercase\expandafter{\romannumeral 5}}-\textit{E} and \textit{F} answer questions (1), (2), and (3) through comparison experiments. Section \textit{\uppercase\expandafter{\romannumeral 5}}-\textit{G} verifies the effectiveness of the proposed basic shape dataset and several designs in our method, which answers question (4).

\subsection{Datasets}

The point cloud classification datasets ModelNet40 \cite{RN424}, ShapeNet55 \cite{RN423}, and ScanObjectNN \cite{RN464} are used in our experiment as the benchmarks. ModelNet40 and ShapNet55 are sets of synthetic objects in point clouds. These datasets are created by collecting CAD models from open-source 3D repositories containing 40 and 55 class categories, respectively. We first introduce ScanObjectNN in the CIL-3D task. It contains 15 categories of real-world point cloud objects from scanned indoor scenes.

Following the split setting in \cite{RN414, RN466, RN467}, ModelNet40 with an increment of 4 classes and ShapeNet55 with 6 classes (7 classes in the last stage) are used for the experiment. Besides, the ScanObjectNN with an increment of 3 classes is introduced as a benchmark.

\subsection{Comparison Methods}

We compare state-of-the-art works in CIL-3D to verify our method's superiority, \textit{i},\textit{e}., I3DOL \cite{RN414}, PACL \cite{RN467}, and InOR-Net \cite{RN466}. Besides, we adapt and apply representative methods in 2D domains to 3D domains for comparisons,  \textit{i},\textit{e}., LwF \cite{RN437}, iCaRL \cite{RN416}, RPS-Net \cite{RN434}, and DGR \cite{RN616}. These above methods mainly rely on exemplars to remember previously learned knowledge. Thus, we introduce the exemplar-free methods, FETRIL \cite{RN428}, EASE \cite{RN421}, MOS \cite{RN611} and SimpleCIL \cite{RN448}, to compare the incremental learning ability without exemplars. Note that for all the pre-training baseline methods, we all pre-train them on our BSA dataset, evaluating which module contributes to the superiority of our method.

\subsection{Implementation Details}

Our method is implemented with Pytorch and PILOT \cite{RN468}, a pre-trained model-based continual learning tool, on a single NVIDIA GeForce RTX 4090 and Intel(R) Xeon(R) Gold 6430 CPU. We follow the dVAE and transformer setups in Point-Bert \cite{RN455} to pre-train our model on the proposed BSA dataset. In the incremental stage, the adapter layers are trained using back-propagation and SGD optimizer with an initial learning rate of 0.01 and batch size of 256.

In the exemplar-free experiment, zero exemplar samples are left in the memory. In the exemplar-based experiment, we follow the same settings as the baseline methods, retaining a fixed number of samples for incremental learning: \textit{M} (exemplar samples) = 800 while learning on ModelNet40, $M \approx 1000$ for ShapeNet55, and $M = 300$ for ScanObjectNN.

We follow iCaRL \cite{RN416} to shuffle class orders with random seed 1993. \textbf{Different seed settings are discussed in section \textit{\uppercase\expandafter{\romannumeral 2}} of the supplementary material.}

\subsection{Evaluation Metrics}

Following the baseline methods \cite{RN414, RN466, RN467} of CIL-3D, we evaluate the classification accuracy ${\mathcal{A}_b}$ in every incremental stage, especially the accuracy ${\mathcal{A}_B}$ in the last stage and the average accuracy ${\bar{\mathcal{A}}}$ along the incremental stage.

\subsection{Comparisons with Exemplar-based Baselines}

\begin{figure*}
    \centering
    \includegraphics[width=1\linewidth]{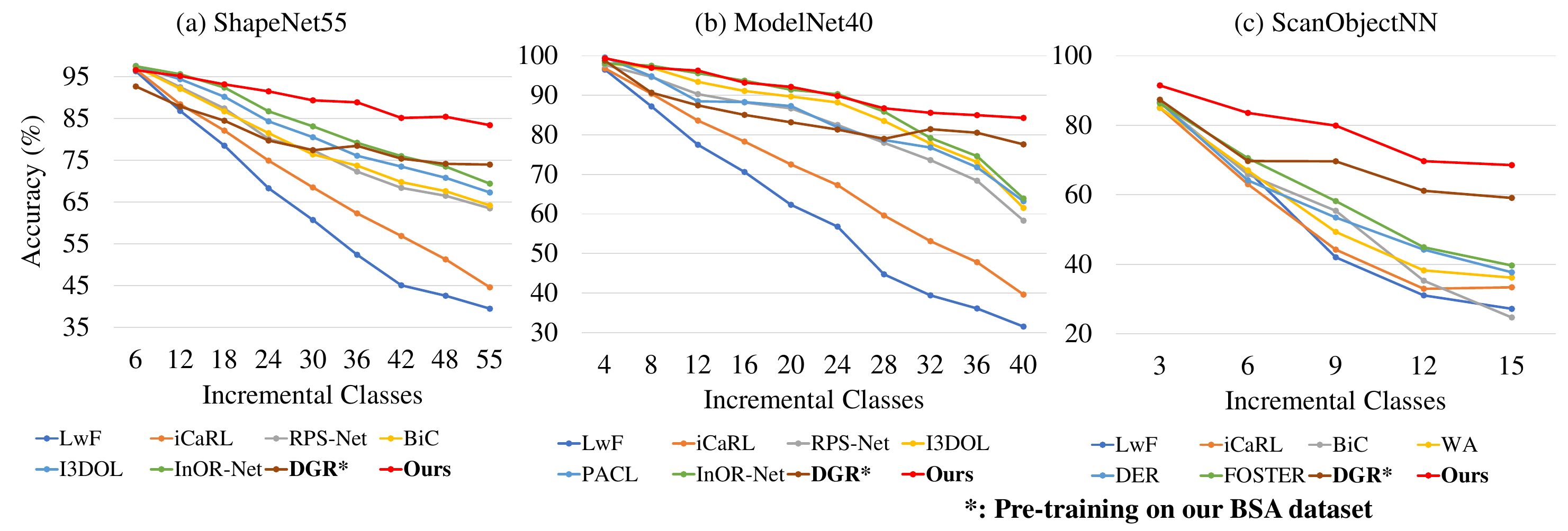}
    \caption{The classification accuracy ${\mathcal{A}_b}$ at each incremental step with different methods on benchmark datasets (exemplar-based comparisons).}
    \label{fig4}
\end{figure*}

Tables \ref{table1} to \ref{table3} list the comparison results on different splitting datasets. Ours obviously outperforms other methods in terms of all metrics. For example, ${\mathcal{A}_B} + 14.0\% $ and $\bar{\mathcal{A}} + 5.6\% $ compared with the state-of-the-art 3D method InOR-Net on ShapeNet55; ${\mathcal{A}_B} + 9.4\% $ and $\bar{\mathcal{A}} + 9.2\% $ compared with the state-of-the-art pre-trained model-based method DGR on ShapeNet55, and figure \ref{fig4} illustrates the classification accuracies of different methods along the class incremental stages.

\begin{table}[H]
    \centering
    \caption{Comparisons on ShapeNet55 with an increment of 6 classes (\textbf{*: Pre-trained Model-based Methods, and we pre-train them on our BSA dataset, the same in Tables \ref{table2} and \ref{table3}}) .}
\begin{tabular}{cccc}
\hline
Method   & \begin{tabular}[c]{@{}c@{}}Exemplars\\ (Samples/Class)\end{tabular} & ${\mathcal{A}_B}$           & ${\bar{\mathcal{A}}}$             \\ \hline
LwF \cite{RN437}   & \multirow{7}{*}{18}& 39.5          & 63.4          \\
iCaRL \cite{RN416}  &                                                                     & 44.6          & 69.5          \\
RPS-Net \cite{RN434}&                                                                     & 63.5          & 78.4          \\
BiC \cite{RN438}   &                                                                     & 64.2          & 78.8          \\
I3DOL  \cite{RN414}  &                                                                     & 67.3          & 81.6          \\
InOR-Net \cite{RN466} &                                                                     & 69.4          & 83.7          \\
 DGR*  \cite{RN616}& & 74.0&80.1\\ \hline
Ours     & 18                                                                  & \textbf{83.4} & \textbf{89.3} \\ \hline
\end{tabular}
    \label{table1}
\end{table}

\begin{table}[H]
    \centering
    \caption{Comparisons on ModelNet40 with an increment of 4 classes.}
\begin{tabular}{cccc}
\hline
Method   & \begin{tabular}[c]{@{}c@{}}Exemplars\\ (Samples/Class)\end{tabular} & ${\mathcal{A}_B}$& ${\bar{\mathcal{A}}}$             \\ \hline
LwF \cite{RN437}   & \multirow{7}{*}{20}& 31.5          & 60.3          \\
iCaRL \cite{RN416}  &                                                                     & 39.6          & 68.9          \\
RPS-Net \cite{RN434}&                                                                     & 58.3          & 81.7          \\
I3DOL  \cite{RN414}   &                                                                     & 61.5          & 85.3          \\
PACL \cite{RN467}  &                                                                     & 63.2          & 83.2         \\
InOR-Net \cite{RN466} &                                                                     & 63.9          & 87.0          \\
 DGR*  \cite{RN616}& & 77.6&84.5\\ \hline
Ours     & 20                                                                  & \textbf{84.3} & \textbf{90.9} \\ \hline
\end{tabular}
    \label{table2}
\end{table}

\begin{table}[H]
    \centering
    \caption{Comparisons on ScanObjectNN with an increment of 3 classes.}
\begin{tabular}{cccc}
\hline
Method   & \begin{tabular}[c]{@{}c@{}}Exemplars\\ (Samples/Class)\end{tabular} & ${\mathcal{A}_B}$           & ${\bar{\mathcal{A}}}$             \\ \hline
LwF \cite{RN437}   & \multirow{7}{*}{20}& 27.2          & 50.4          \\
iCaRL \cite{RN416}  &                                                                     & 33.4         & 51.7          \\
BiC \cite{RN438} &                                                                     & 24.7          & 53.5          \\
WA \cite{RN447}   &                                                                     & 36.2          & 55.2          \\
DER \cite{RN417}  &                                                                     & 37.7          & 57.2         \\
FOSTER \cite{RN431} &                                                                     & 39.7          & 59.9          \\
 DGR*  \cite{RN616}& & 59.1&69.4\\ \hline
Ours     & 20                                                                  & \textbf{68.6} & \textbf{78.6} \\ \hline
\end{tabular}
    \label{table3}
\end{table}

Results on different datasets reflect similar experimental phenomena. Specifically, the methods migrated from 2D-CIL, including LwF, iCaRL, RPS-Net, and BiC, do not work well in the incremental learning of 3D point clouds. These methods relieve the model forgetting through specific designs to some extent. However, they are still incapable of remembering the increasingly diverse 3D geometry features along with the growing label space. The CIL-3D methods (I3DOL, InOR-Net, and PACL) characterize the irregular point clouds and help the models remember the unique characters of different 3D Classes. These methods resist the model forgetting by a large margin. However, the class representations are easily confused while the label space grows. Compared to DGR, which underwent the same pre-training as ours, our results still significantly outperform, demonstrating that our method continues to exhibit superiority despite the pre-training dataset.

Our BSA dataset contains abundant and meaningful 3D geometry features in local regions. The backbone learns the contexts between the local geometries, widely recognizing the relationships between the geometry contexts and the class representations. Based on this, our method regularizes the object representation of the same class category and widens the gap between different ones. Thus, our method remembers the geometry features well and learns the class representations well.

\begin{figure*}
    \centering
    \includegraphics[width=1\linewidth]{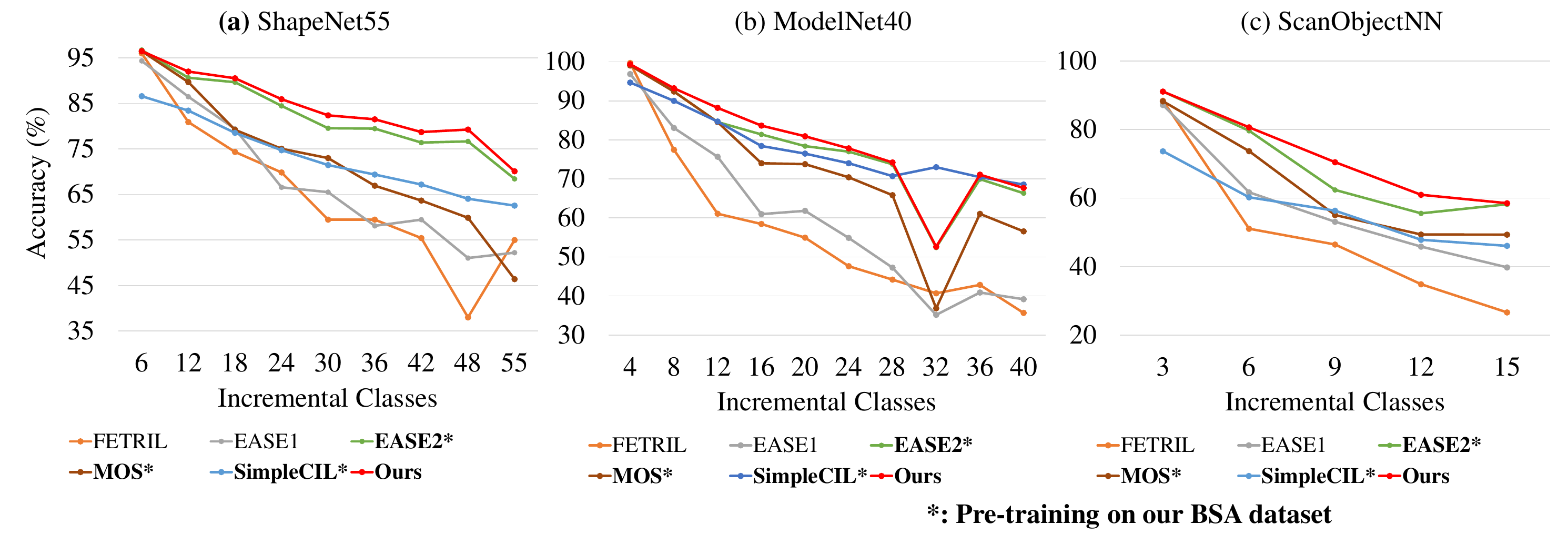}
    \caption{The classification accuracy ${\mathcal{A}_b}$ at each incremental step with different methods on benchmark datasets  (exemplar-free comparisons).}
    \label{fig5}
\end{figure*}

\begin{table}
    \centering
    \caption{Comparisons on different datasets with zero exemplar samples (\textbf{*: Pre-trained Model-based Methods, and we pre-train them on our BSA dataset, leaving only EASE1 without pre-training for comparisons}).}
    \begin{tabular}{ccccccc}
\hline
\multirow{2}{*}{Method}           & \multicolumn{2}{c}{ShapeNet55} \rule{0pt}{10pt}& \multicolumn{2}{c}{ModelNet40} & \multicolumn{2}{c}{ScanObjectNN} \\ \cline{2-7} 
                                  & ${\mathcal{A}_B}$ \rule{0pt}{10pt}& ${\bar{\mathcal{A}}}$             & ${\mathcal{A}_B}$& ${\bar{\mathcal{A}}}$             & ${\mathcal{A}_B}$& ${\bar{\mathcal{A}}}$          \\ \hline
PointCLIMB  \cite{RN411}                     & -              & -             & 8.2            & 30.1          & -               & -              \\
FETRIL \cite{RN428}                           & 55.0           & 65.4          & 35.7           & 56.3          & 26.6            & 49.5           \\
EASE$_1$* \cite{RN421}& 52.2           & 68.1          & 39.2           & 59.6          & 39.8            & 57.5           \\
EASE$_2$* \cite{RN421}& 68.4           & 82.4          & 66.3           & 77.6          & 58.2            & 69.4           \\
 MOS* \cite{RN611}& 46.4& 72.3& 56.6& 71.4& 49.3&63.1\\
 SimpleCIL* \cite{RN448}& 62.6& 73.1& \textbf{68.6}& 78.1& 46.0&56.8\\ \hline
Ours                              & \textbf{70.1}  & \textbf{84.1} & 67.7& \textbf{78.9} & \textbf{58.5}     & \textbf{72.3}    \\ \hline
\end{tabular}
    \label{table4}
\end{table}
\subsection{Comparison with Exemplar-free Baselines}

Table \ref{table4} lists the comparisons on different datasets with zero exemplar samples, and figure \ref{fig5} illustrates the details in different learning stages. Except for PointCLIMB, the only method explored exemplar-free CIL-3D, all methods were originally proposed in the image-based CIL, and we used them in CIL-3D for comparisons. Compared with PointCLIMB, our method outperforms it by a large margin (ModelNet40: ${\mathcal{A}_B} + 59.5\% $ and $\bar{\mathcal{A}} + 48.8\% $). In comparison to state-of-the-art methods based on pre-trained models, our approach consistently outperforms nearly all evaluation metrics on every dataset.

The exemplar-free experimental setting is more challenging than the exemplar-based one. Thus, the results on all the datasets degrade a lot. PointCLIMB employs an optimization process where a teacher model is initially trained on base classes, and a student model handles incremental novel tasks by copying the teacher's weights. However, the base classes contain biased knowledge, and the relief of model forgetting is insignificant. FETRIL proposed a simple yet effective translation method of class features and improved incremental learning performance.

We also conduct experiments to verify whether our basic shape dataset helps the pre-trained model-based 2D methods work well in the 3D domains. We introduce EASE \cite{RN421}, MOS \cite{RN611}, and SimpleCIL \cite{RN448}, replacing the image embedding layer with the point cloud one for CIL-3D. On the one hand, the accuracy boost is noticeable while pre-training EASE$_2$ on the basic shapes (ShapeNet55: $\bar{\mathcal{A}} + 14.3\% $ compared with EASE$_1$ without pre-training). It proves the effectiveness of our proposed basic shape dataset. On the other, EASA$_2$, MOS, and SimpleCIL work worse than our proposed method, even pre-training on the same dataset. Our method better uses the basic shape knowledge and demonstrates a more reliable ability for incremental learning.

\subsection{Ablation Studies}

As the cornerstone of our methodology, we demonstrate the effectiveness of the \textit{BSA pre-training} through a comparative analysis with LLM and further corroborate its validity in a specifically designed scenario. Besides, we validate the effectiveness of the \textit{exemplar-adaptive regularization}, which is a core design of our CIL-3D framework.

We assess the influence of the \textit{self-supervised learning strategy in pre-training} by comparing it with alternative approaches. We also verify that the \textit{adapters} are useful by comparing the experimental results with or without them. Although these two designs are not the innovative aspects of our work, we prove that our choices are well-justified.

\subsubsection{Effectiveness of the BSA pre-training (verified by comparing LLMs)}

We replace our pre-training model with PointLLMs to compare the experimental results. PointLLM \cite{RN601} (\textit{ECCV} 2024, Best Paper Candidate) integrates LLMs (Large Language Models) with point cloud data to enable advanced 3D understanding and semantic interaction. Table \ref{table5} illustrates the comparisons using baselines with different model sizes and embedding dims.

\begin{table}[H]
    \centering
    \caption{Comparisons between LLMs and ours. 13B and 7B denote model sizes of PointLLM; PointBert v1.2 and v1.2 indicate the point encoder with different embedding dims.}
\begin{tabular}{ccc}
\hline
Method   & ${\mathcal{A}_B}$ \rule{0pt}{10pt}           & ${\bar{\mathcal{A}}}$             \\ \hline
PointLLM-13B\_PointBert-v1.1& 29.1& 51.8\\
PointLLM-13B\_PointBert-v1.2& 42.1& 62.3
\\
PointLLM-7B\_PointBert-v1.1& 30.0& 51.9
\\
 PointLLM-13B\_PointBert-v1.2& 41.1&61.8
\\ \hline
Ours     & \textbf{67.7}& \textbf{78.9}\\ \hline
\end{tabular}
    \label{table5}
\end{table}

Despite the extensive resources required for training LLMs on massive datasets, our zero-collection-cost pre-training framework achieves superior performance by leveraging the inherent geometric richness of our BSA datasets. While LLMs focus on global feature modeling, often overlooking local geometric details, our approach explicitly captures these fine-grained patterns, enabling precise differentiation between object categories and delivering SOTA results in CIL.

\subsubsection{Effectiveness of the BSA pre-training (verified in a specially designed scenario)}
We conduct the following ablation experiment to evaluate the performance difference between the zero-collection-cost basic shapes and the collected dataset.

As discussed in the introduction, the existing classification datasets in point clouds retain a small domain gap. We cannot pick one for pre-training and another one for continual learning. To address this, we mix ShapeNet55 with ModelNet40, assigning the objects of similar semantics with the same class label and forming a novel dataset of 71 class categories. We pick 26 categories of objects in the mixed dataset as the pre-training dataset (\textit{Mix-Pre}.) and use the remaining 45 categories (\textit{Mix-CIL}) for continual learning. Thus, the pre-training dataset retains a large domain gap with the CIL one.

We pre-train the same backbone model on the Mix-Pre. dataset and our proposed basic shapes, respectively. Incremental learnings on the Mix-CIL dataset, with an increment of 5 classes, are carried out. Table \ref{table6} illustrates the CIL results with different pre-trained models.

\begin{table}[H]
    \centering
    \caption{Experimental results on Mix-CIL with different pre-training datasets.}
\begin{tabular}{ccc}
\hline
 \begin{tabular}[c]{@{}c@{}}Pre-training\\ Dataset\end{tabular} & ${\mathcal{A}_B}$  & $\bar{\mathcal{A}} $   \\ \hline
 Mix-Pre.                                                       & 71.4 & 84.6 \\
                       Basic Shapes (Ours)                                            & 71.3 & 83.7 \\ \hline
\end{tabular}
    \label{table6}
\end{table}

Interestingly, even with two different pre-training datasets, the methods demonstrate similar continual learning abilities. It indicates that our zero-collection-cost dataset provides the same sufficient and meaningful basic geometry knowledge as the collected dataset.

\begin{figure*}
    \centering
    \includegraphics[width=0.75\linewidth]{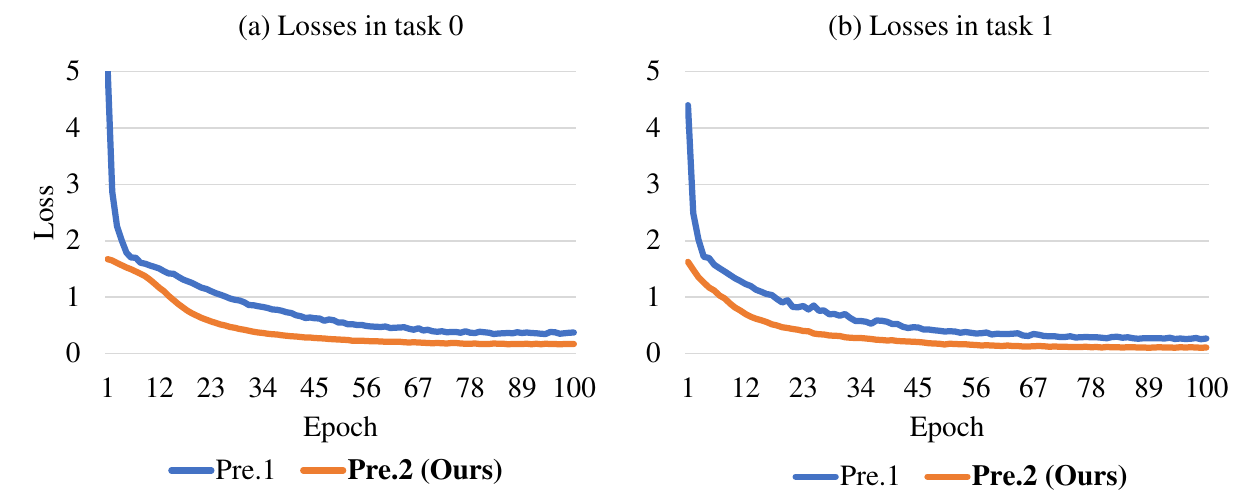}
    \caption{Losses along the training epochs in different incremental tasks. Pre. 1 denotes pre-training on Mix-Pre. ; Pre. 2 denotes pre-training on the basic shapes. The model pre-trained with our basic shape dataset converges faster than that with the collected dataset Mix-Pre.}
    \label{fig6}
\end{figure*}

Moreover, as shown in figure \ref{fig6}, our basic shape dataset promotes model convergence in the incremental learning stage. It provides the model with fundamental knowledge that is easier to generalize than the collected datasets.

Last but not least, the Mix-Pre. dataset is designed to keep a big gap with the Mix-CIL; it only works in specially designed scenarios. Differently, our dataset retains a big domain gap with every collected dataset; it can work in every scenario.

\subsubsection{Effectiveness of the exemplar-adaptive regularizations}

We remove the exemplar-adaptive regularization item in the prototype calculation to verify its effectiveness. Table \ref{table7} illustrates the experimental results with or without regularizations, considering exemplar-free and (-based) settings. The regularization helps improve the experimental results. It proves that the regularization item resists the  model's catastrophic forgetting.

\begin{table}[H]
    \centering
    \caption{Experimental results on ModelNet40 with or without regularizations, considering different exemplar settings.}
    \begin{tabular}{cccc}
\hline
Exemplar             & \begin{tabular}[c]{@{}c@{}}Regularization\\ Item\end{tabular} & ${\mathcal{A}_B}$            & $\bar{\mathcal{A}} $           \\ \hline
\multirow{2}{*}{w/o}  & w/o             & 66.3 & 77.2 \\
                     & w/            & \textbf{67.7}          & \textbf{78.9}          \\ \hline
\multirow{2}{*}{w/} & w/o             & 80.1          & 87.2 \\
                     & w/            & \textbf{84.3} & \textbf{90.9}          \\ \hline
\end{tabular}
    
    \label{table7}
\end{table}

\subsubsection{Effectiveness of the self-supervised learning strategy in pre-training}

Point-BERT and PointMAE \cite{RN610} employ two representative strategies for the self-supervised pre-training: Point-BERT uses a BERT-style approach, masking random point cloud regions and predicting them based on context, while PointMAE employs a masked autoencoding strategy, focusing on reconstructing the full point cloud from heavily masked inputs, emphasizing geometric recovery.

\begin{table}[H]
    \centering
    \caption{Experimental results with different self-supervised learning strategies on different datasets.}
    \begin{tabular}{cccc}
\hline
Dataset             & \begin{tabular}[c]{@{}c@{}}Self-supervised\\learning strategy\end{tabular} & ${\mathcal{A}_B}$            & $\bar{\mathcal{A}} $           \\ \hline
\multirow{2}{*}{ModelNet40}  & MAE-style             & 63.5 & 76.8 \\
                     & BERT-style            & \textbf{67.7}          & \textbf{78.9}          \\ \hline
\multirow{2}{*}{ScanObjectNN} & MAE-style             & 47.3          & 68.0 \\
                     & BERT-style            & \textbf{58.5} & \textbf{72.3}          \\ \hline
\end{tabular}
    
    \label{table8}
\end{table}

Table \ref{table8} shows MAE-style self-supervised pre-training results in better CILs than the BERT- ones, especially for the complicated ScanObject dataset. PointMAE's focus on geometric reconstruction limits its semantic understanding and task adaptability, making it less effective than Point-BERT in continual learning, as Point-BERT's contextual modeling better captures high-level relationships and generalizes across diverse tasks.

\subsubsection{Effectiveness of the adapters}

A frozen backbone with increasing adapter layers balances the CIL performance and the model size well, which has been proven in MOS \cite{RN611} and EASE \cite{RN421}. The method with fixed parameters may work well on some CIL datasets (SimpleCIL on ModelNet in table \ref{table4} as an example). However, for a CIL dataset that significantly differs from the pre-trained dataset, such as the real-world ScanobjectNN, which has a large domain gap compared to the synthetic BSA, it becomes challenging to adapt the learned knowledge to new tasks without an adapter, resulting in poor experimental results (DGR and SimpleCIL on ScanobjectNN in tables \ref{table3} and \ref{table4} as examples).

\begin{table}[H]
    \centering
    \caption{Comparisons between our method with or without adapter dropouts.}
\begin{tabular}{ccc}
\hline
 \begin{tabular}[c]{@{}c@{}}Adapter\\Dropout\end{tabular} & ${\mathcal{A}_B}$  & $\bar{\mathcal{A}} $   \\ \hline
$\checkmark$                                                       & 57.2 & 72.4 \\
                                                                   & \textbf{67.7} & \textbf{78.9} \\ \hline
\end{tabular}
    \label{table9}
\end{table}

We supplementary experiments to compare the CIL performance of a model initialized with a larger size (2×DGR with 180.1M Params.) and our method (90.5M with 4.7M Params. increments per task); our method with a much smaller model still outperforms the 2×DGR obviously (ShapeNet55: ${\mathcal{A}_B} + 2.9\% $ and $\bar{\mathcal{A}} + 3.3\% $; ModelNet40: ${\mathcal{A}_B} + 1.4\% $ and $\bar{\mathcal{A}} + 1.2\% $; ScanobjectNN: ${\mathcal{A}_B} + 6.2\% $ and $\bar{\mathcal{A}} + 5.2\% $). It also verifies the effectiveness of the increasing adapters.

Besides, we conduct ablations to verify the frozen adapters' effectiveness. Table \ref{table9} illustrates that dropping some adapters out results in performance degradation, proving that even in tasks with novel trainable adapters, the frozen adapters still play important roles in revisiting the former category knowledge.

\section{Conclusion}

This paper explores the class-incremental learning problem in 3D point clouds. It is the first work addressing the CIL-3D problem using a pre-trained method. We propose a basic shape dataset and overcome the lack of a pre-training dataset in 3D point clouds with zero collection cost. We also propose a framework embedded with geometry knowledge, resisting the catastrophic forgetting of models. Experiments prove that our methods work well, outperforming other baseline methods largely. Besides, it helps the model converge quickly. We think this work is a good benchmark for CIL-3D with pre-trained methods; the proposed basic shape dataset can help lots of downstream applications.

\bibliographystyle{ieeetr}

\begin{thebibliography}{10}

\bibitem{RN414}
J.~Dong, Y.~Cong, G.~Sun, B.~Ma, and L.~Wang, ``I3dol: Incremental 3d object learning without catastrophic forgetting,'' in {\em AAAI Conference on Artificial Intelligence}, pp.~6066--6074, 2021.

\bibitem{RN466}
J.~Dong, Y.~Cong, G.~Sun, L.~Wang, L.~Lyu, J.~Li, and E.~Konukoglu, ``Inor-net: Incremental 3-d object recognition network for point cloud representation,'' {\em IEEE Transactions on Neural Networks and Learning Systems}, vol.~34, no.~10, pp.~6955--6967, 2023.

\bibitem{RN410}
Y.~Liu, Y.~Cong, G.~Sun, T.~Zhang, J.~Dong, and H.~Liu, ``L3doc: Lifelong 3d object classification,'' {\em IEEE Transactions on Image Processing}, vol.~30, pp.~7486--7498, 2021.

\bibitem{RN413}
M.~Zamorski, M.~Stypulkowski, K.~Karanowski, T.~Trzcinski, and M.~Zieba, ``Continual learning on 3d point clouds with random compressed rehearsal,'' {\em Computer Vision and Image Understanding}, vol.~228, p.~103621, 2023.

\bibitem{RN452}
H.~Shin, J.~K. Lee, J.~Kim, and J.~Kim, ``Continual learning with deep generative replay,'' in {\em Advances in Neural Information Processing Systems}, pp.~2990--2999, 2017.

\bibitem{RN411}
S.~Kundargi, T.~Anvekar, R.~A. Tabib, and U.~Mudenagudi, ``Pointclimb: An exemplar-free point cloud class incremental benchmark,'' {\em arXiv}, vol.~abs/2304.06775, 2023.

\bibitem{RN465}
Z.~Wang, Z.~Zhang, C.-Y. Lee, H.~Zhang, R.~Sun, X.~Ren, G.~Su, V.~Perot, J.~G. Dy, and T.~Pfister, ``Learning to prompt for continual learning,'' in {\em IEEE Conference on Computer Vision and Pattern Recognition}, pp.~139--149, 2022.

\bibitem{RNA460}
Z.~Wang, Z.~Zhang, S.~Ebrahimi, R.~Sun, H.~Zhang, C.~Lee, X.~Ren, G.~Su, V.~Perot, J.~G. Dy, and T.~Pfister, ``Dualprompt: Complementary prompting for rehearsal-free continual learning,'' in {\em European Conference on Computer Vision}, vol.~13686 of {\em Lecture Notes in Computer Science}, pp.~631--648, Springer, 2022.

\bibitem{RN421}
D.-W. Zhou, H.-L. Sun, H.-J. Ye, and D.-C. Zhan, ``Expandable subspace ensemble for pre-trained model-based class-incremental learning,'' in {\em IEEE Conference on Computer Vision and Pattern Recognition}, pp.~23554--23564, 2024.

\bibitem{RN617}
X.~Cao, H.~Lu, X.~Liu, and M.-M. Cheng, ``Class incremental learning for image classification with out-of-distribution task identification,'' {\em IEEE Transactions on Multimedia}, pp.~1--14, 2025.

\bibitem{RN425}
J.~Deng, W.~Dong, R.~Socher, L.-J. Li, K.~Li, and L.~Fei-Fei, ``Imagenet: A large-scale hierarchical image database,'' in {\em IEEE Conference on Computer Vision and Pattern Recognition}, pp.~248--255, 2009.

\bibitem{RN426}
Y.~Zhang, Z.~Yin, J.~Shao, and Z.~Liu, ``Benchmarking omni-vision representation through the lens of visual realms,'' in {\em European Conference on Computer Vision}, pp.~594--611, 2022.

\bibitem{RN423}
A.~X. Chang, T.~A. Funkhouser, L.~J. Guibas, P.~Hanrahan, Q.-X. Huang, Z.~Li, S.~Savarese, M.~Savva, S.~Song, H.~Su, J.~Xiao, L.~Yi, and F.~Yu, ``Shapenet: An information-rich 3d model repository,'' {\em arXiv}, vol.~abs/1512.03012, 2015.

\bibitem{RN424}
Z.~Wu, S.~Song, A.~Khosla, F.~Yu, L.~Zhang, X.~Tang, and J.~Xiao, ``3d shapenets: A deep representation for volumetric shapes,'' in {\em IEEE Conference on Computer Vision and Pattern Recognition}, pp.~1912--1920, 2015.

\bibitem{RN601}
R.~Xu, X.~Wang, T.~Wang, Y.~Chen, J.~Pang, and D.~Lin, ``Pointllm: Empowering large language models to understand point clouds,'' in {\em European Conference on Computer Vision}, pp.~131--147.

\bibitem{RN604}
T.~L. Heath, {\em The Thirteen Books of the Elements, Vol. 3}, vol.~3.
\newblock Courier Corporation, 2013.

\bibitem{RN605}
D.~Hilbert and S.~Cohn-Vossen, {\em Geometry and the Imagination}, vol.~87.
\newblock American Mathematical Soc., 2021.

\bibitem{RN609}
F.~D. Ching, {\em Architecture: Form, space, and order}.
\newblock John Wiley \& Sons, 2023.

\bibitem{RN607}
D.~Marr, {\em Vision: A computational investigation into the human representation and processing of visual information}.
\newblock MIT press, 2010.

\bibitem{RN449}
D.-W. Zhou, Q.-W. Wang, Z.-H. Qi, H.-J. Ye, D.-C. Zhan, and Z.~Liu, ``Class-incremental learning: A survey,'' {\em arXiv}, vol.~abs/2302.03648, 2024.

\bibitem{RN459}
F.~Zhu, X.-Y. Zhang, C.~Wang, F.~Yin, and C.-L. Liu, ``Prototype augmentation and self-supervision for incremental learning,'' in {\em IEEE Conference on Computer Vision and Pattern Recognition}, pp.~5871--5880, 2021.

\bibitem{RN429}
Y.~Liu, B.~Schiele, and Q.~Sun, ``Rmm: Reinforced memory management for class-incremental learning,'' in {\em Advances in Neural Information Processing Systems}, pp.~3478--3490, 2021.

\bibitem{RN451}
H.~Zhao, H.~Wang, Y.~Fu, F.~Wu, and X.~Li, ``Memory-efficient class-incremental learning for image classification,'' {\em IEEE Transactions on Neural Networks and Learning Systems}, vol.~33, no.~10, pp.~5966--5977, 2022.

\bibitem{RN450}
Y.~Liu, Y.~Su, A.-A. Liu, B.~Schiele, and Q.~Sun, ``Mnemonics training: Multi-class incremental learning without forgetting,'' in {\em IEEE Conference on Computer Vision and Pattern Recognition}, pp.~12242--12251, 2020.

\bibitem{RN620}
Y.~Cui, W.~Deng, X.~Xu, Z.~Liu, Z.~Liu, M.~Pietikainen, and L.~Liu, ``Uncertainty-guided semi-supervised few-shot class-incremental learning with knowledge distillation,'' {\em IEEE Transactions on Multimedia}, vol.~25, pp.~6422--6435, 2023.

\bibitem{RN622}
Y.~Luo, R.~Cong, X.~Liu, H.~H.~S. Ip, and S.~Kwong, ``Modeling inner- and cross-task contrastive relations for continual image classification,'' {\em IEEE Transactions on Multimedia}, vol.~26, pp.~10842--10853, 2024.

\bibitem{RN437}
Z.~Li and D.~Hoiem, ``Learning without forgetting,'' in {\em European Conference on Computer Vision}, pp.~614--629, 2016.

\bibitem{RN416}
S.-A. Rebuffi, A.~Kolesnikov, G.~Sperl, and C.~H. Lampert, ``icarl: Incremental classifier and representation learning,'' in {\em IEEE Conference on Computer Vision and Pattern Recognition}, pp.~5533--5542, 2017.

\bibitem{RN438}
Y.~Wu, Y.~Chen, L.~Wang, Y.~Ye, Z.~Liu, Y.~Guo, and Y.~Fu, ``Large scale incremental learning,'' in {\em IEEE Conference on Computer Vision and Pattern Recognition}, pp.~374--382, 2019.

\bibitem{RN446}
E.~Belouadah and A.~Popescu, ``Il2m: Class incremental learning with dual memory,'' in {\em IEEE International Conference on Computer Vision}, pp.~583--592, 2019.

\bibitem{RN439}
D.-W. Zhou, H.-J. Ye, and D.-C. Zhan, ``Co-transport for class-incremental learning,'' in {\em ACM International Conference on Multimedia}, pp.~1645--1654, 2021.

\bibitem{RN440}
C.~Simon, P.~Koniusz, and M.~Harandi, ``On learning the geodesic path for incremental learning,'' in {\em IEEE Conference on Computer Vision and Pattern Recognition}, pp.~1591--1600, 2021.

\bibitem{RN441}
A.~Douillard, M.~Cord, C.~Ollion, T.~Robert, and E.~Valle, ``Podnet: Pooled outputs distillation for small-tasks incremental learning,'' in {\em European Conference on Computer Vision}, pp.~86--102, 2020.

\bibitem{RN442}
X.~Hu, K.~Tang, C.~Miao, X.-S. Hua, and H.~Zhang, ``Distilling causal effect of data in class-incremental learning,'' in {\em IEEE Conference on Computer Vision and Pattern Recognition}, pp.~3957--3966, 2021.

\bibitem{RN443}
F.~Zhu, Z.~Cheng, X.-Y. Zhang, and C.-L. Liu, ``Class-incremental learning via dual augmentation,'' in {\em Advances in Neural Information Processing Systems}, pp.~14306--14318, 2021.

\bibitem{RN430}
D.~Lopez-Paz and M.~Ranzato, ``Gradient episodic memory for continual learning,'' in {\em Advances in Neural Information Processing Systems}, pp.~6467--6476, 2017.

\bibitem{RN436}
J.~Kirkpatricka, R.~Pascanu, N.~Rabinowitz, J.~Veness, G.~Desjardins, A.~A. Rusu, K.~Milan, J.~Quan, T.~Ramalho, and A.~Grabska-Barwinska, ``Overcoming catastrophic forgetting in neural networks,'' in {\em Proceedings of the National Academy of Sciences of the United States of America.}, 2017.

\bibitem{RN444}
S.~Hou, X.~Pan, C.~C. Loy, Z.~Wang, and D.~Lin, ``Learning a unified classifier incrementally via rebalancing,'' in {\em IEEE Conference on Computer Vision and Pattern Recognition}, pp.~831--839, 2019.

\bibitem{RN445}
F.~M. Castro, M.~J. Marin-Jimenez, N.~Guil, C.~Schmid, and K.~Alahari, ``End-to-end incremental learning,'' in {\em European Conference on Computer Vision}, pp.~241--257, 2018.

\bibitem{RN447}
B.~Zhao, X.~Xiao, G.~Gan, B.~Zhang, and S.-T. Xia, ``Maintaining discrimination and fairness in class incremental learning,'' in {\em IEEE Conference on Computer Vision and Pattern Recognition}, pp.~13205--13214, 2020.

\bibitem{RN417}
S.~Yan, J.~Xie, and X.~He, ``Der: Dynamically expandable representation for class incremental learning,'' in {\em IEEE Conference on Computer Vision and Pattern Recognition}, pp.~3014--3023, 2021.

\bibitem{RN431}
F.-Y. Wang, D.-W. Zhou, H.-J. Ye, and D.-C. Zhan, ``Foster: Feature boosting and compression for class-incremental learning,'' in {\em European Conference on Computer Vision}, pp.~398--414, 2022.

\bibitem{RN432}
D.-W. Zhou, Q.-W. Wang, H.-J. Ye, and D.-C. Zhan, ``A model or 603 exemplars: Towards memory-efficient class-incremental learning,'' in {\em International Conference on Learning Representations}, 2023.

\bibitem{RNA461}
O.~Ostapenko, M.~M. Puscas, T.~Klein, P.~J{\"{a}}hnichen, and M.~Nabi, ``Learning to remember: {A} synaptic plasticity driven framework for continual learning,'' in {\em IEEE Conference on Computer Vision and Pattern Recognition}, pp.~11321--11329, 2019.

\bibitem{RN434}
J.~Rajasegaran, M.~Hayat, S.~H. Khan, F.~S. Khan, and L.~Shao, ``Random path selection for continual learning,'' in {\em Advances in Neural Information Processing Systems}, pp.~12648--12658, 2019.

\bibitem{RN621}
Y.~Li, W.~Cao, W.~Xie, J.~Li, and E.~Benetos, ``Few-shot class-incremental audio classification using dynamically expanded classifier with self-attention modified prototypes,'' {\em IEEE Transactions on Multimedia}, vol.~26, pp.~1346--1360, 2024.

\bibitem{RN448}
D.-W. Zhou, Z.-W. Cai, H.-J. Ye, D.-C. Zhan, and Z.~Liu, ``Revisiting class-incremental learning with pre-trained models: Generalizability and adaptivity are all you need,'' {\em International Journal of Computer Vision}, 2024.

\bibitem{RN453}
J.~Devlin, M.-W. Chang, K.~Lee, and K.~Toutanova, ``Bert: Pre-training of deep bidirectional transformers for language understanding,'' in {\em Annual Conference of the North American Chapter of the Association for Computational Linguistics: Human Language Technologies}, pp.~4171--4186, 2019.

\bibitem{RN454}
H.~Bao, L.~Dong, S.~Piao, and F.~Wei, ``Beit: Bert pre-training of image transformers,'' in {\em International Conference on Learning Representations}, 2022.

\bibitem{RN455}
X.~Yu, L.~Tang, Y.~Rao, T.~Huang, J.~Zhou, and J.~Lu, ``Point-bert: Pre-training 3d point cloud transformers with masked point modeling,'' in {\em IEEE Conference on Computer Vision and Pattern Recognition}, pp.~19291--19300, 2022.

\bibitem{RN46}
C.~R. Qi, H.~Su, K.~Mo, and L.~J. Guibas, ``Pointnet: Deep learning on point sets for 3d classification and segmentation,'' in {\em IEEE Conference on Computer Vision and Pattern Recognition}, pp.~77--85, 2017.

\bibitem{RN469}
J.~T. Rolfe, ``Discrete variational autoencoders,'' in {\em International Conference on Learning Representations}, 2017.

\bibitem{RN458}
N.~Houlsby, A.~Giurgiu, S.~Jastrzebski, B.~Morrone, Q.~d. Laroussilhe, A.~Gesmundo, M.~Attariyan, and S.~Gelly, ``Parameter-efficient transfer learning for nlp,'' in {\em International Conference on Machine Learning}, pp.~2790--2799, 2019.

\bibitem{RN464}
M.~A. Uy, Q.-H. Pham, B.-S. Hua, D.~T. Nguyen, and S.-K. Yeung, ``Revisiting point cloud classification: A new benchmark dataset and classification model on real-world data,'' in {\em IEEE International Conference on Computer Vision}, pp.~1588--1597, 2019.

\bibitem{RN467}
X.~Wang and X.~Wei, ``Continual learning for pose-agnostic object recognition in 3d point clouds,'' {\em arXiv}, vol.~abs/2209.04840, 2022.

\bibitem{RN616}
J.~He, ``Gradient reweighting: Towards imbalanced class-incremental learning,'' in {\em IEEE Conference on Computer Vision and Pattern Recognition}, pp.~16668--16677.

\bibitem{RN428}
G.~Petit, A.~Popescu, H.~Schindler, D.~Picard, and B.~Delezoide, ``Fetril: Feature translation for exemplar-free class-incremental learning,'' in {\em IEEE Winter Conference on Applications of Computer Vision}, pp.~3900--3909, 2023.

\bibitem{RN611}
H.-L. Sun, D.~Zhou, H.~Zhao, L.~Gan, D.-C. Zhan, and H.-J. Ye, ``Mos: Model surgery for pre-trained model-based class-incremental learning,'' in {\em AAAI Conference on Artificial Intelligence}.

\bibitem{RN468}
LAMDA. \url{https://github.com/sun-hailong/LAMDA-PILOT}.

\bibitem{RN610}
Y.~Pang, W.~Wang, F.~E.~H. Tay, W.~Liu, Y.~Tian, and L.~Yuan, ``Masked autoencoders for point cloud self-supervised learning,'' in {\em European Conference on Computer Vision}, pp.~604--621.
\end{thebibliography}

\end{document}